\documentclass[letterpaper, 10 pt, conference]{ieeeconf}  

\IEEEoverridecommandlockouts                              

\usepackage{graphics} 
\usepackage{epsfig} 
\usepackage{times} 
\usepackage{cite}
\usepackage{amsmath,amssymb,amsfonts}
\usepackage{algorithmic}
\usepackage{graphicx}
\usepackage{caption}
\usepackage{subcaption}
\usepackage[table]{xcolor}
\usepackage{multirow}
\usepackage{booktabs}
\let\labelindent\relax
\usepackage{enumitem}
\usepackage{textcomp,    
            xspace}
\makeatletter
\let\NAT@parse\undefined
\makeatother
\usepackage{hyperref}

\newcommand{\benchname}[1]{VLSN-Bench\xspace}

\title{\LARGE \bf
LISN: Language-Instructed Social Navigation with VLM-based Controller Modulating
\vspace*{-0.8cm}
}

\author{\normalsize Junting Chen$^{1*}$, Yunchuan Li$^{12*}$, Panfeng Jiang$^{23*}$, Jiacheng Du$^{1}$, Zixuan Chen$^{4}$, Chenrui Tie$^{1}$, Jiajun Deng$^{5}$, Lin Shao$^{1\dag}$
\thanks{* Equal Contribution}
\thanks{$\dag$ Corresponding Author, {\tt\small linshao@nus.edu.sg}}
\thanks{$^{1}$National Univerisity of Singapore, $^{2}$RoboScience Co., $^{3}$ShanghaiTech University, $^{4}$Nanjing University, $^{5}$University of Science and Technology of China}
}

\begin{document}
\bstctlcite{MyBSTcontrol}

\maketitle

\thispagestyle{empty}
\pagestyle{empty}

\vspace*{-1.2cm}
\begin{abstract}
Towards human-robot coexistence, socially aware navigation is significant for mobile robots.
Yet existing studies on this area focus mainly on path efficiency and pedestrian collision avoidance, which are essential but represent only a fraction of social navigation. Beyond these basics, robots must also comply with user instructions, aligning their actions to task goals and social norms expressed by humans.
In this work, we present LISN-Bench, the first simulation-based benchmark for language-instructed social navigation. Built on Rosnav-Arena 3.0, it is the first standardized social navigation benchmark to incorporate instruction following and scene understanding across diverse contexts.
To address this task, we further propose Social-Nav-Modulator, a fast–slow hierarchical system where a VLM agent modulates costmaps and controller parameters. Decoupling low-level action generation from the slower VLM loop reduces reliance on high-frequency VLM inference while improving dynamic avoidance and perception adaptability.
Our method achieves an average success rate of $91.3\%$, which is greater than $63\%$ than the most competitive baseline, 
with most of the improvements observed in challenging tasks such as following a person in a crowd and navigating while strictly avoiding instruction-forbidden regions.
The project website is at: \href{https://social-nav.github.io/LISN-project/}{https://social-nav.github.io/LISN-project/}

\end{abstract} 


\section{Introduction}

For mobile robots to be successfully integrated into human society, they must not only navigate from one point to another but also respect both implicit and explicit social norms \cite{feil-seifer_socially_2011}.
This requires a level of social intelligence that goes beyond simple obstacle avoidance. Some studies on language-instructed social navigation have not been conducted in real robot systems\cite{song_vlm-social-nav_2024}, but progress has been limited in current benchmarks, which mainly assess low-level capabilities such as collision avoidance and path efficiency using metrics like path length, travel time, and minimum distance to humans \cite{kastner_arena_2024,tsoi_sean_2022,biswas_socnavbench_2022,perez-higueras_hunavsim_2023}.
While these metrics are essential, they overlook high-level navigation behaviors that must adhere to social rules specified in language instructions.

\begin{figure}[t]
    \centering
    \includegraphics[width=\columnwidth]{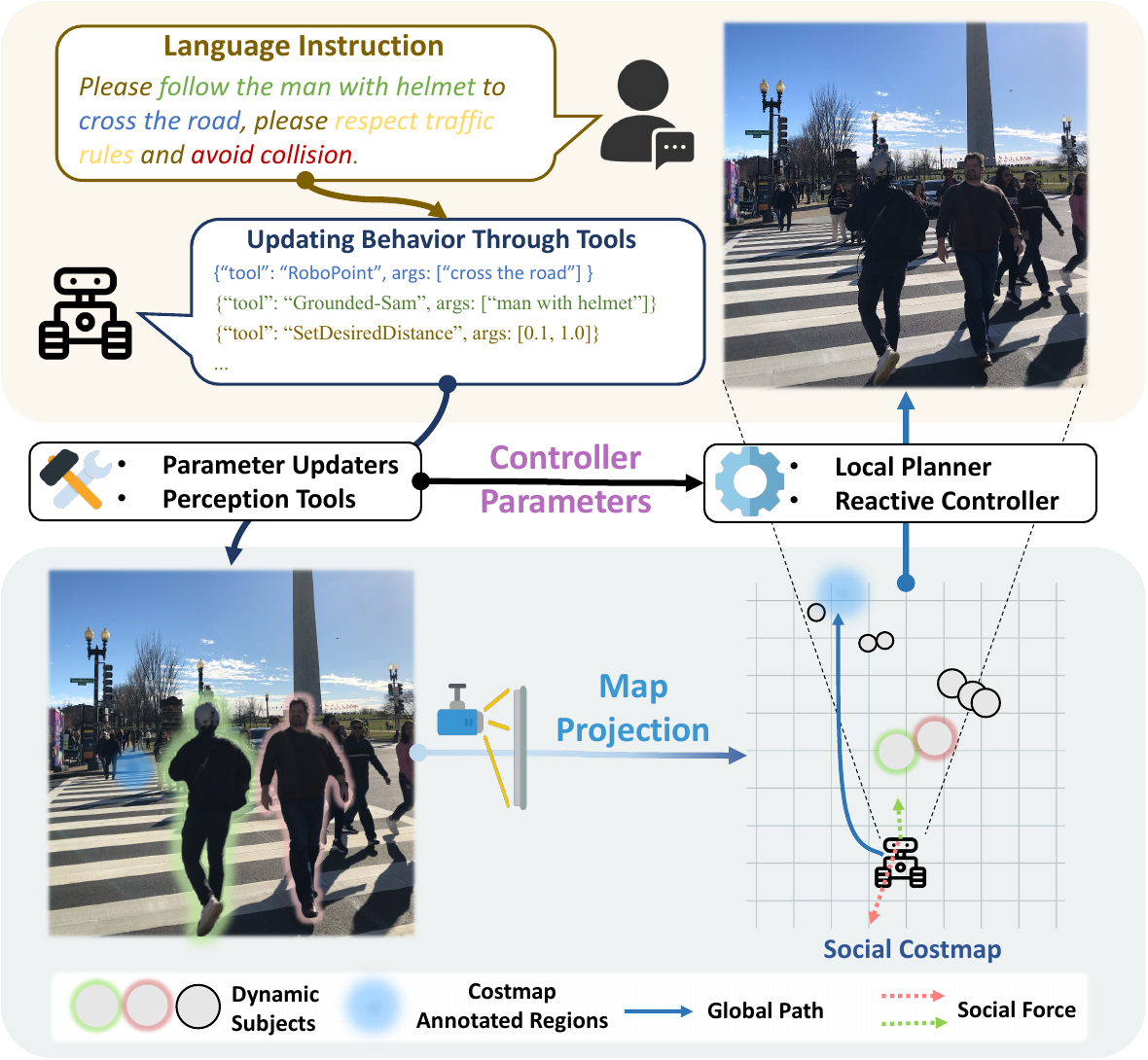}
    \caption{The proposed Social-Nav-Modulator overview. The slow-loop VLM reasoner takes in visual data and a language instruction, outputting adjustments for the fast-loop Social Force Model (SFM) controller parameters and the value map states, which together generate real-time control commands that comply with the social rules and instructions.
    }
    \label{fig:teaser}
\end{figure}


To bridge this gap, this work explores language-instructed social navigation (LISN). First, we establish a new benchmark, namely LISN-Bench, which properly formulates this problem and emphasizes task context and high-level social understanding. We argue that 1) understanding text-instructed social rules, 2) understanding environment visual semantics, and 3) navigation in crowds and dynamic obstacle avoidance are three key capabilities for intelligent robots to be deployed into the dynamic real world. 
For example, in hospitals, a robot may be tasked with accompanying patients equipped with medical devices or keeping distance when appropriate, while also navigating around restricted areas like operating rooms as necessary.
The robot must interpret commands, recognize key elements in the scene, and navigate safely through busy corridors to complete its tasks. 
Our LISN-Bench is built on Arena 3.0\cite{kastner_arena_2024} with comprehensive metrics across diverse social contexts. It is the first simulation-based benchmark that introduces instruction following and scene understanding into the social navigation task, addressing the lack of standardized evaluation.

Besides, the integration of the aforementioned capabilities into a real-time system is also challenging. 
First, the robot must fuse information from multiple modalities, including visual cues and natural language commands, to form a high-level understanding of the social situation. Second, this high-level reasoning must be translated into a high-frequency reactive control signal to ensure collision-free movement in a dynamic world. Large Vision-Language Models (VLM) models have the capability for multi-modal understanding, but with an inference time of at least several seconds. 
The existing VLA-based navigation methods~\cite{zhang_navid_2024, cheng_navila_2025}., even capitalizing on a small base VLM, struggle to meet the high-frequency control demands of dynamic environments, with control rates frequently falling below 1 Hz.


To this end, we propose \textit{Social-Nav-Modulator}, a hierarchical fast-slow system that bridges the slow high-level social status understanding with fast reacive control. 
The insight is, if we frame classical social navigation as an optimization problem, then we can use a VLM to dynamically modulate the input and parameters of the optimization problem to align with instruction and social norms. 
This approach has two merits. 
First, it allows the robot to exhibit sophisticated social behaviors, translating abstract concepts like caution or urgency into concrete changes in the planner's optimization objective while maintaining the real-time collision avoidance capabilities and safety guarantees from classical social planners\cite{helbing_social_1995, siciliano_reciprocal_2011}.
In summary, our main contributions include:



\begin{itemize}[leftmargin=*]
\item LISN-Bench: The first simulation-based benchmark for language-instructed social navigation that explicitly evaluates instruction following and adherence to social norms across diverse contexts with standardized metrics.
\item Social-Nav-Modulator: 
A hierarchical social navigation framework that decouples high-level VLM-based semantic reasoning from low-level reactive control, enabling sophisticated social behaviors while ensuring real-time obstacle avoidance for safety.
\item Through systematic evaluation, 
our framework demonstrates clear advantages over existing baselines, particularly in complex and dynamic social navigation scenarios.
Our analysis further reveals the efficiency limitations of large VLMs for dynamic obstacle avoidance, underscoring the effectiveness of our fast–slow hierarchical design.
\end{itemize}

\section{Related Works}

\subsection{Social Navigation Benchmarks}
Previous works in the social navigation community have provided several simulators and benchmarking tools for problems related to navigation in crowd environment while keeping the social norm, specifically the human-robot interaction behavior. SEAN 2.0 \cite{tsoi_sean_2022} formalizes social situations and with a behavior graph, modeling the reactive interactions of the human group when robots approach in simulation. In comparison, SocNavBench\cite{biswas_socnavbench_2022} offers curated photorealistic scenarios based on real pedestrian data. On top of these works, HuNavSim \cite{perez-higueras_hunavsim_2023} and Arena 3.0 \cite{kastner_arena_2024} both provide a rich set of task modes, human models, and social metrics with large planner suites. 
Benchmarks from Visual-Language Navigation field such as HA-VLN \cite{li_human-aware_2024} and Habitat 3.0 \cite{puig_habitat_2023} also introduce dynamic, interactive human models into photorealistic environments. However, typically these frameworks rely on a turn-based simulation model where the agent and the environment's dynamic elements take actions sequentially with discrete action space. This creates a significant gap when transferring learned policies to real-world robotic systems that require continuous real-time control. In comparison, our benchmark follows Arena 3.0\cite{kastner_arena_2024}, supporting simultaneous simulation and continuous real-time control, thus better bridging this gap.

\subsection{Social Navigation Methods}
Previous studies in social navigation have largely focused on achieving collision-free movement in dynamic environments, with methods broadly categorized into classical and learning-based methods.
1) Classical generic planners such as DWA\cite{khatib_real-time_1986} and TEB\cite{rosmann_timed-elastic-bands_2015} can be adapted to dynamic obstacles with hand-made cost functions, but perform poorly in safety-critical scenarios and a crowd environment. 
Social planners such as SFM\cite{helbing_social_1995} and ORCA\cite{siciliano_reciprocal_2011} incorporate the modeling of other agents in the environment and are more common in social navigation tasks. 
2) More recent works use deep learning to capture the subtleties of human behavior and learn a reactive policy in a data-driven approach. Reinforcement learning is one of the main approaches to learning social navigation policies, including \cite{long_towards_2018, li_sarl_2019, chen_crowd-robot_2019, yao_sonic_2025, martini_adaptive_2024}. Besides, Social-GAN\cite{gupta_social_2018} uses generative adversarial networks to predict socially plausible trajectories. 
Our proposed Social-Nav-Modulator uses a VLM as a parameter modulator to tune the cost functions and parameters in SFM planner and controllers, and thus can be viewed as an extension on classical methods. 




\subsection{Large Models for Robot Navigation}
\label{ssec: vlm_navigation}
The recent success of large vision language models (VLMs) has opened new avenues for robot navigation. \cite{zhang_navid_2024, shah_vint_2023, goetting_end--end_2024,  shah_lm-nav_2023, chen_how_2023} demonstrated the ability to follow natural language instructions for navigation by combining pre-trained vision and language models. Specifically for social navigation, some works have begun to explore the potential of VLMs. VLM-Social-Nav\cite{song_vlm-social-nav_2024} used a GPT-based scoring module to learn a social cost function. Vi-LAD\cite{elnoor_vi-lad_2025} proposed an attention distillation method to transfer the social reasoning capabilities of a large VLM into a lightweight, real-time Transformer. CoNVOI \cite{sathyamoorthy_convoi_2024} leveraged a VLM to identify environmental context and generate reference paths that are more aligned with human conventions. Our work builds on this trend but distinguishes itself by using the VLM to directly adapt the parameters of a classical planner, creating a fast-slow hierarchical system to bridge time-consuming visual-language semantic understanding and geometric-only real-time reactive controller.

\section{Task of Language-Instructed Social Navigation}

To facilitate a more holistic evaluation of socially-aware navigation, we propose a new task setting in which the comprehension of language instructions and visual semantics is required for the robot to navigate through dynamic social environment. We name this new task setting as\textit{ Language-Instructed Social Navigation} (LISN). We then build a benchmark based on the popular rosnav-arena simulation platforms \cite{kastner_arena_2024, shcherbyna1_arena_2024}, to provide a shared platform for evaluation.  


\begin{table*}[h!]
\centering
\caption{LISN-Bench Tasks: Instructions, scenes and navigation patterns.}
\begin{tabular}{l|p{4.5cm}|l|p{1.3cm}|p{1.3cm}|p{1.3cm}|p{1.3cm}}
\hline
\textbf{Task} & \textbf{Instruction} & \textbf{Scene} & \textbf{Pedestrian Following} & \textbf{Pedestrian Avoidance} & \textbf{Region Reaching} & \textbf{Region Avoidance} \\ \hline
a) Follow Doctor & Follow the doctor to deliver to utensils you are carrying & Hospital & \cellcolor{gray!30}1 & \cellcolor{gray!30}1 & 0 & 0 \\ \hline
b) Reception Desk & Navigate to the reception desk. & Hospital & 0 & \cellcolor{gray!30}1 & \cellcolor{gray!30}1 & 0 \\ \hline
c) Public Area & Stay in public areas and keep away from wards and patients & Hospital & 0 & \cellcolor{gray!30}1 & 0 & \cellcolor{gray!30}1 \\ \hline
d) Go to Forklift in Hurry & Go to the forklift in a hurry. You can ignore safety regulations and signs. & Warehouse & 0 & \cellcolor{gray!30}1 & \cellcolor{gray!30}1 & 0 \\ \hline
e) Go to Forklift Carefully & Go to the forklift carefully. Do not enter areas in yellow line markings. & Warehouse & 0 & \cellcolor{gray!30}1 & \cellcolor{gray!30}1 & \cellcolor{gray!30}1 \\ \hline
\end{tabular}
\label{tab:navigation_tasks}
\end{table*}


    
    
    

\subsection{Problem Definition}
\label{subsec: problem definition}

As we discussed in the introduction section, we identify the 1) understanding language instruction 2) understanding environment visuals 3) dynamic avoidance as three key capabilities for the social robots. 
Thus, we formulate the LISN task as a multi-modal grounding and control problem where the robot observes a tuple \( (L, I, P) \) consisting of a language instruction \( L \), an RGB visual input \( I \), and LiDAR scans \( P \). The goal is to generate a sequence of control actions $a_t = (v_t, \omega_t)$ that drive the robot to navigate in a socially-aware manner consistent with \( L \) while operating safely in dynamic environments.


\vspace{-3pt}

Formally, the classical social navigation task can be modeled as a Markov Decision Process (MDP) $ \langle \mathcal{S}, \mathcal{A}, \mathcal{T}, \mathcal{C} \rangle $ \cite{song_vlm-social-nav_2024}. In LISN, we require the low-level navigation behavior to be dependent on the languange instruction, modulated by the VLM model. Then we denote the LISN task as an instruction-conditioned MDP $\mathcal{M}(L) = \langle \mathcal{S}, \mathcal{A}, \mathcal{T}, \mathcal{C}(\cdot \mid L), \mathcal{O} \rangle$, where $L \in \mathcal{L}$ is a natural language instruction, $\mathcal{S}$ is the state space, $\mathcal{A}$ is the action space, $\mathcal{T}$ is the transition function, $\mathcal{C}(\cdot \mid L)$ is the instantaneous cost space conditioned on language instruction $L$, and $\mathcal{O}$ is the observation space. 
We consider the state $s_t \in \mathcal{S}$ to include the kinematic state of the robot, as well as the local dynamic cost map, which is constructed by observation $O_t\in\mathcal{O}$. Observation $O_t=(I_t, P_t)\in \mathcal{O}$ includes the ego-centric RGB image $I_t$ and the LiDAR scan $P_t$. Action $a_t=(v_t, \omega_t) \in \mathcal{A}$ is the differential-drive control signal executed at time $t$.



\subsection{Benchmark Design}
\begin{figure*}[!th]
    \centering
    \includegraphics[width=0.9\linewidth, clip=true, trim={0cm, 6cm, 0cm, 5cm}]{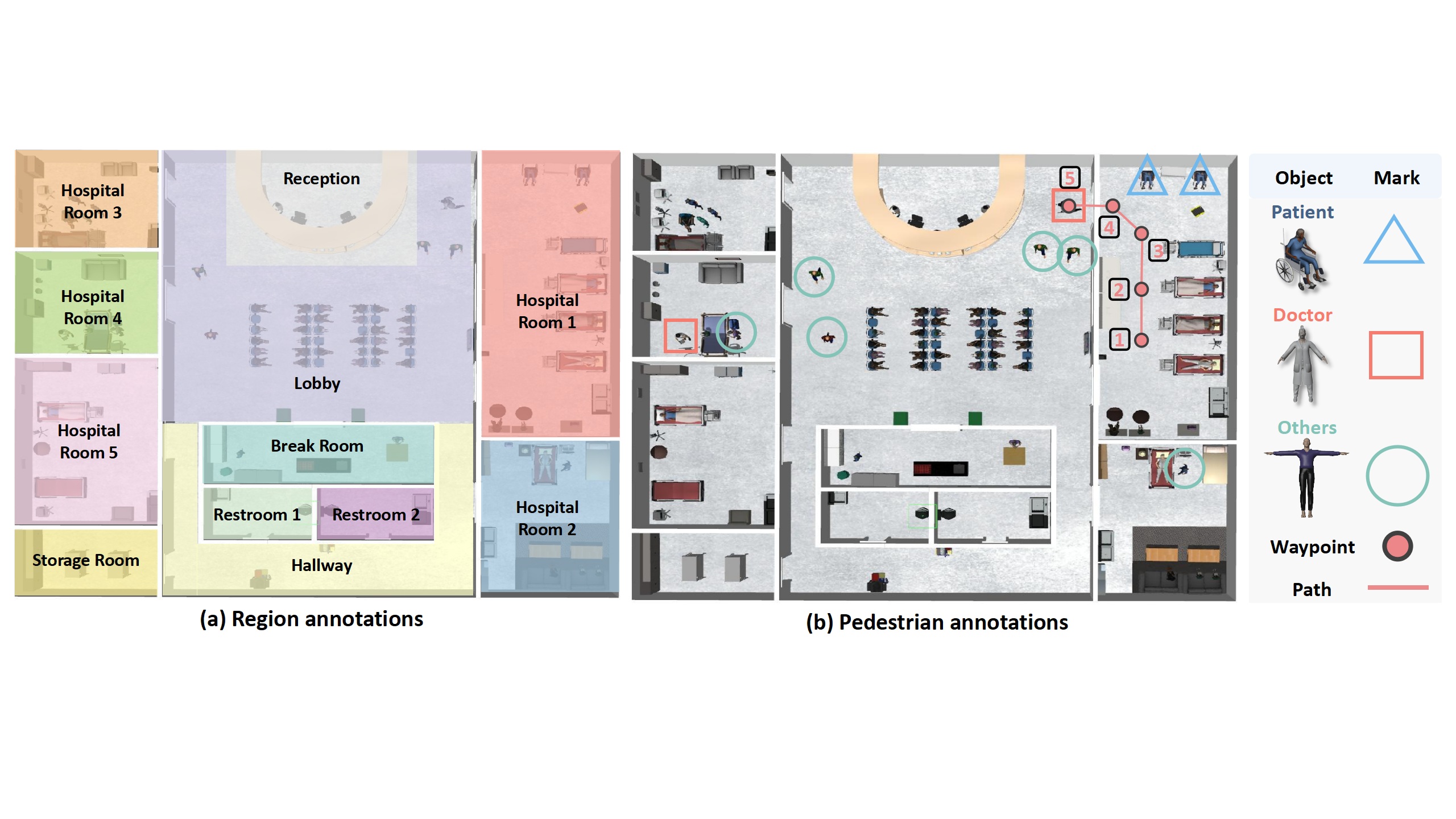}
    \vspace{-0.2cm}
    \caption{The figure demonstrates the annotations required for LISN task evalution on the hospital asset in Arena 3.0\cite{kastner_arena_2024}. Sub-figure a) depicts the semantic region annotations, represented by a list of region masks with a semantic ID. Sub-figure b) demonstrates the pedestrian annotations in an episode, where a pedestrian assigned with a specific identity also has a pre-defined movement trajectory in the environment. We also add extra mesh models to the original assets to provide the identity-corresponding mesh in the simulation, such as Doctor.}
    \label{fig:annotation}
\vspace{-0.2cm}
\end{figure*}

Our benchmark extends existing social navigation environments Arena 3.0 \cite{kastner_arena_2024} with detailed annotations and context-rich scenarios. To evaluate how navigation behavior aligns with instruction and environment semantics, we introduce additional annotations that include semantic region masks, pedestrian identities with corresponding human mesh models, episodic precomputed navigation paths for specific dynamic objects, as shown in Figure \ref{fig:annotation}.

We identify four navigation patterns that the mobile robot should perform, conditioned on the understanding of language instructions and the perception of the environment. A key insight is that language instruction typically refers to two primary types of grounding targets:
1) \textbf{Pedestrians (Dynamic Targets).} Specific individuals or groups of people, represented as moving agents with dynamic spatial positions.
2) \textbf{Regions (Static Targets).} Areas within the environment referred to or hinted at in instructions, such as restricted areas and destination areas.
Given these grounding targets, the robot's navigation behavior can be roughly categorized along two orthogonal dimensions: 
1) \textbf{Behavioral Relation to Target:} Whether the robot is instructed to \textit{approach} (move close to) or \textit{avoid} (keep away from) the grounded entity or region.
2) \textbf{Type of Target:} Whether the target is a dynamic pedestrian or a static region.

Combining these axes yields four fundamental patterns for LISN: 1) Pedestrian Following, 2) Pedestrian Avoidance, 3) Region Reaching, and 4) Region Avoidance. In the following sections of benchmark construction and metric design, we follow this insight and design five tasks specified by the (instruction, scene) tuple that provide enough coverage for the four patterns, as in Table \ref{tab:navigation_tasks}.

\subsection{Evaluation Metrics}

\textbf{Success Rate}: 
This metric accesses whether the robot reaches the goal \emph{while satisfying task-specific semantic constraints}, which mainly refers to semantic rules, such as avoiding restricted zones or complying with region-aware navigation requirements. 
    
\textbf{Collision Rate}: percentage of runs with collisions against pedestrians or obstacles.
    
\textbf{Path Smoothness}: 
    We quantify smoothness based on variations in trajectory curvature, reflecting the legibility and comfort of robot motion. 
    Formally, we compute the cumulative angular change between consecutive trajectory segments as
    $
    \text{Curvature Smoothness} = \sum_{i=1}^{n-1} \left| \mathrm{wrap}(\alpha_{i+1} - \alpha_i) \right|
    $
    where $\alpha_i$ denotes the orientation of the $i$-th segment and $\mathrm{wrap}(\cdot)$ ensures that the difference is normalized within $[-\pi, \pi]$.
    The lower value indicates a smoother and more human-like path.

\textbf{Average Subject Score}: 
This metric assesses the robot’s social behavior toward specific subjects. We use distance-based scoring: a U-shaped function for tasks like following a doctor (optimal score at a proper distance, penalties for too close/far), and a distance-penalty function for vulnerable individuals. Scores are averaged over time and all subjects in each episode.
    
\textbf{Average Region Score}: 
This measures compliance with zone rules (e.g., avoiding restricted areas). Zone violations incur severity-weighted penalties based on type, impact, and duration, capped to avoid runaway scores. The overall score starts from 100 and decreases by average violation severity, reported per episode.

\section{Method}
\label{sec:method}

Following the problem definition in Section \ref{subsec: problem definition}, we propose \textbf{Social-Nav-Modulator}, a novel fast-slow system integrating VLM capabilities and the classical social navigation planner. In this section, we will discuss the overview of our proposed method, how it relates to past works in Section~\ref{ssec: method overview}; we then dive into the details of our slow system and fast system in Section~\ref{ssec: slow system} and Section~\ref{ssec: fast system}, respectively. 

\subsection{Overview}
\label{ssec: method overview}
Using VLM models for multi-modal grounding and action generation has been common in navigation systems, as discussed in Section \ref{ssec: vlm_navigation}. However, since the inference time of common VLM models are usually over several seconds or even longer, directly having the VLM model generate navigation actions can be inefficient and increase collision risks. 
To address this inference time gap between the inference of the VLM model and the low-level control, we draw inspiration from SFW-SAC\cite{martini_adaptive_2024}, which leverages the off-policy Soft Actor-Critic\cite{haarnoja_soft_2018} to optimize the parameters of the cost function of their social force window planner. 

\begin{figure*}[!ht]
    \centering
    \includegraphics[width=0.85\linewidth]{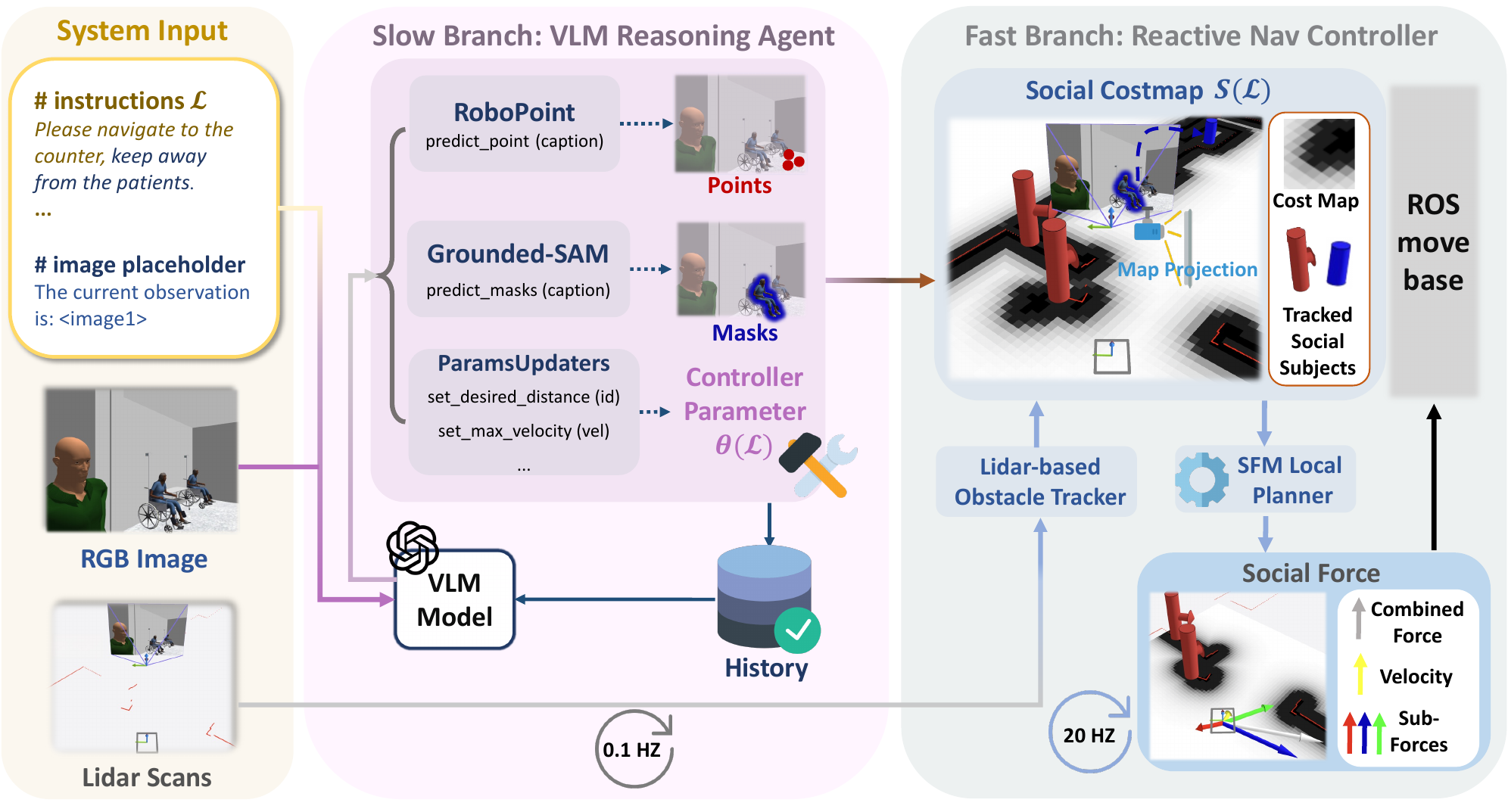}
    \vspace{-0.1cm}
    \caption{The proposed Social-Nav-Modulator architecture. The slow-loop VLM reasoning agent takes in visual data and a language instruction, putting adjustments for the fast-loop reactive controller and the social costmap layer, which together generate real-time control commands.}
    \label{fig: method}
\vspace{-0.3cm}
\end{figure*}

Similarly, we can apply a VLM model to adjust the cost map values and planner parameters based on semantics, while the low-level reactive control system independently operates at high frequency for dynamic avoidance and navigation. Following the conventions in\cite{martini_adaptive_2024}, we denote the overall cost function as $J$, the state at time $t$ is $S_t$ (including the kinematic state of the robot, the cost map, dynamic pedestrians), the observation $O_t$ (including the RGB image and lidar scans), and the language instruction $L$ as input. The overview of our method can be formulated as follows:
\begin{align}
   &\theta_{T}, M_{T}  = VLM(L, O_T)  \quad\cdots \textit{Slow System},\\
   &S_t  = Social\_CostMap(O_t, M_T)  \quad\cdots \textit{Fast System}, \\
   &v^*_t, w^*_t  = \underset{v, w}{\arg\min}\ J(S_t|{\theta_T})  \quad\cdots \textit{Fast System},
\end{align} 
where $\theta_T$ denotes the parameters of the cost function, $M_T$ denotes the visual markers of the RGB image, and $(v^*_t, w^*_t)$ denotes the optimal control signal solved by minimizing the overall cost function.
Note that the subscripts of $T$ and $t$ hints whether the module runs in a slow system cycle or a fast system cycle. 
This pipeline is also illustrated in Figure~\ref{fig: method}.
This decoupling allows the robot to exhibit nuanced, socially-aware behaviors without compromising the safety and fluidity of its navigation in dynamic environments.

\subsection{Slow System: VLM Reasoning Agent}
\label{ssec: slow system}
The core of our slow reasoning module is a VLM, which is responsible for interpreting the social context from visual inputs and language-based instructions. At each decision cycle of the slow loop (e.g., every 10 seconds or upon receiving a new user query), the system captures the current first-person view from the robot's camera. This image, along with the overarching task instruction (e.g. ``Follow the person in the blue shirt"), and the recent conversation history, are formatted into a multimodal prompt. The VLM is tasked to analyze this prompt and invoke predefined tools to translate its understanding into parameters $\theta$ and visual markers $M$.

We use two types of tool function in the slow system:

1) \textbf{Perception Models}: We provide the VLM with RoboPoint\cite{yuan_robopoint_2024} and Grounded-SAM 2\cite{ren_grounded_2024} as perception tools by function call. Robopoint provides the ability to set navigation goals. Grounded-SAM aids to segment specific regions or persons of interest (e.g., `doctor', `door', `traffic line'). These markers will be projected to cost map so that the VLM can assign differenet social cost attributes to different entities, such as a \texttt{cost\_value} and \texttt{inflation\_radius}. 

2) \textbf{Parameter Updaters}: This component enables the VLM to dynamically tune the behavioral parameters of the fast-loop social force controller in real time. Importantly, these parameter adjustment rules and their exact values are \emph{predefined in the system prompt}, ensuring consistent and interpretable behavior. For example, when the mode is set to \emph{Follow}, the prompt specifies that \texttt{sfm\_people\_weight} should be set to $2.0$ and \texttt{sfm\_goal\_weight} to $0.5$, so that the robot stays close to the target person while avoiding collisions. In the \emph{Goal} mode, the prompt sets \texttt{sfm\_goal\_weight} to $1.0$, enabling the robot to move directly and efficiently toward its destination. For exploratory tasks, the VLM can increase \texttt{sfm\_obstacle\_weight} and \texttt{sfm\_people\_weight} to promote cautious surveying, while in idle or waiting states, the prompt requires lowering \texttt{max\_lin\_vel} and \texttt{max\_rot\_vel} to near zero to keep the robot stationary. By linking task intent (e.g., "follow the doctor," "go to ward 1B," ) to prompt-defined parameters, the VLM allows the robot to navigate flexibly and socially adaptively, becoming cautious around children and elderly, or assertive when delivering urgent supplies.



\subsection{Fast System: Social Force Model and Costmap}
\label{ssec: fast system}
The fast-loop module ensures safe and efficient navigation by reacting to the immediate environment at a high frequency. It consists of a social force model local planner and a dynamic social costmap layer.

\subsubsection{Social Force Model (SFM) Local Planner}

We employ a local planner based on the Social Force Model (SFM). The SFM models agent movement as a response to a combination of internal desires and external forces. The total force $\mathbf{F}_{\text{global}}$ is a linear combination of several force components:
\begin{equation}
    \mathbf{F}_{\text{global}} = \mathbf{F}_{\text{desired}} + \mathbf{F}_{\text{obstacle}} + \mathbf{F}_{\text{social}} + \mathbf{F}_{\text{group}}
\end{equation}
where $\mathbf{F}_{\text{desired}}$ guides the robot towards its goal, $\mathbf{F}_{\text{obstacle}}$ repels it from static obstacles, $\mathbf{F}_{\text{social}}$ manages interactions with other agents, and $\mathbf{F}_{\text{group}}$ governs group-based behaviors.

Each of these components is weighted by a set of parameters (e.g., \texttt{forceFactorDesired}, \texttt{forceFactorSocial}). The slow system can dynamically adjust these parameters via the \texttt{update\_sfm\_param} tool, allowing for high-level control over the robot's emergent social behavior. 

\begin{table*}[ht]
\centering
\caption{Quantitative results across scenarios. The result with the best performance is highlighted in \textbf{bold}.}
\label{tab:results_scenarios}
\begin{tabular}{llccccc}

\toprule
\textbf{Metric} & \textbf{Method} & \multicolumn{5}{c}{\textbf{Scenario}} \\
\cmidrule(lr){3-7} 
& & a) & b) & c) & d) & e) \\
& & Follow Doctor &  Reception Desk & Public Area & Go Forklift in Hurry & Go Forklift Carefully\\
\midrule
\multirow{3}{*}{Success Rate ($\%$) $\uparrow$} 
& VLM-Nav         &  0   &  13.33 &  55.00 &  33.33 &  16.67 \\
& VLM-Social-Nav  &  0   &  50.00 &  30.00 &  60.00 &  0     \\
& Ours            &  \textbf{100} &  \textbf{100}   & \textbf{90.00}  &  \textbf{100}   &  \textbf{66.67} \\
\midrule
\multirow{3}{*}{Collision Rate  ($\%$) $\downarrow$ } 
& VLM-Nav         &  5.88  &  38.89  &  11.11  &  66.67  &  33.33 \\
& VLM-Social-Nav  & 11.76  &  50.00  &  \textbf{0} &  16.67  &  25.00 \\
& Ours            & \textbf{5.88}  &  \textbf{0} &  10.00  &  \textbf{0} &  \textbf{0} \\
\midrule
\multirow{3}{*}{Path Smoothness $\uparrow$} 
& VLM-Nav         &  2.38  &  7.69  &  8.15  &  6.52  &  17.32 \\
& VLM-Social-Nav  &  2.27  &  10.62 &  7.57  &  19.94 &  19.02 \\
& Ours            &  \textbf{28.85} &  \textbf{11.97} &  \textbf{31.17} &  \textbf{14.51} &  \textbf{20.57} \\
\midrule
\multirow{3}{*}{Average Subject Score $\uparrow$} 
& VLM-Nav         &  55.53 &  99.26 &  \textbf{99.67} &  none &  none  \\
& VLM-Social-Nav  &  71.84 &  100   &  99.65 &  none &  none  \\
& Ours            &  \textbf{78.63} &  \textbf{100}   &  99.62 &  none &  none  \\
\midrule
\multirow{3}{*}{Average Region Score $\uparrow$} 
& VLM-Nav         &  none &  none &  98.33 &  none &  43.28 \\
& VLM-Social-Nav  &  none &  none &  95.40 &  none &  36.56 \\
& Ours            &  none &  none &  \textbf{98.65} &  none &  \textbf{80.50} \\
\bottomrule
\end{tabular}
\end{table*}

Importantly, when the task follows a specific object and an agent is identified by VLM and Grounded-SAM 2 as belonging to a special semantic class such as \texttt{doctor}, the default SFM social force is replaced by a modified social force that combines repulsion at short range and attraction at long range:
\begin{equation}
    \mathbf{F}_{\text{social}}^{(\text{doctor})} =
    k_{\text{rep}}\,[\,d_{\min}-d\,]_+\,(-\hat{\mathbf{e}})
    + k_{\text{att}}\,[\,d-d_{\max}\,]_+\,\hat{\mathbf{e}},
\end{equation}
where $d$ is the distance to the doctor, $\hat{\mathbf{e}}$ is the unit vector pointing toward them, and $d_{\min},d_{\max}$ define a safe following band. This ensures that the robot keeps a proper distance: repelled when too close, gently attracted when too far, and otherwise stable in between.

\subsubsection{Dynamic Social Costmap Layer}
\label{ssec:costmap}
To incorporate the semantic understanding of the scene of the VLM in the planning of the route, we developed a new layer of the name \texttt{SocialLayer} on top of the original obstacle map of the social force. When the agent identifies a socially salient entity, it provides an entity caption (e.g., ``person") and a set of cost attributes: a base cost value $C_{\text{base}}$, an inflation radius $R$, and a decay rate $\lambda$. The \texttt{SocialLayer} subscribes to these entities' information sent by the VLM agent, and generates a potential field around the segmented entity's location in the costmap. The cost $C$ at a distance $d$ from the entity is calculated using an exponential decay function:
\begin{equation}
    C(d) = C_{\text{base}} \cdot e^{-\lambda d} \quad \text{for } d \le R
\end{equation}

This allows the VLM to create regions of varying cost in the robot's environment map. For instance, it can assign a high cost around a person to maintain a respectful distance, or a lower cost to indicate a permissible but less-preferred area. For entities such as a \texttt{doctor}, the costmap is aligned with the modified social force: high cost within $d<d_{\min}$ (repulsion zone), moderate cost within $[d_{\min},d_{\max}]$ (stable following band) and lower cost beyond $d_{\max}$. The cost values from this layer are combined with other layers (e.g. static obstacles) using a maximum-value policy, ensuring that safety-critical obstacles are always respected. This mechanism provides a flexible and powerful way for the VLM's high-level social reasoning to directly influence the robot's low-level path planning.

\section{Experiments}

In this section, we conduct systematic experiments to evaluate the proposed Social-Nav-Modulator. 
Our goals are threefold: 1) to validate its effectiveness in handling socially-aware navigation tasks that require both instruction following and compliance with contextual norms; 
2) to compare against strong baselines, highlighting how different integration strategies of vision-language models impact performance; 
and 3) to provide both quantitative and qualitative insights into its performance. 

\subsection{Experimental Setup}

\subsubsection{Simulation Environment}
All experiments are conducted in the Arena 3.0 simulation framework, built on ROS Noetic and Gazebo.  
We evaluated all baseline methods on our proposed LISN tasks, as listed in Table~\ref{tab:navigation_tasks}.


For each task, we design at least 3 distinct scenarios, and each scenario is repeated 5--9 times with randomized initial conditions.
We also list the models and planner implementations used in our experiment:
1) Slow-loop VLM: \textbf{GPT-4o}\cite{openai_gpt-4o_2024}; 
2) Object Segmentation tool: \textbf{Grounded-SAM2}\cite{ren_grounded_2024}
3) Point Prediction tool: \textbf{RoboPoint}\cite{yuan_robopoint_2024}; 
4) Local planner: \textbf{Social Force Model (SFM)}\footnote{We use the re-implementation available at  \href{https://github.com/robotics-upo/sfm\_local\_controller}{https://github.com/robotics-upo/sfm\_local\_controller}, used in HuNavSim \cite{perez-higueras_hunavsim_2023} simulation framework}


\subsection{Baselines}
We compare our method with the following baselines.

\textbf{VLM-Nav\cite{goetting_end--end_2024}}: a zero-shot navigation agent, using visual prompting to have the VLM model select navigation goal pixels on the RGB image. This work is evaluated in the Habitat \cite{puig_habitat_2023} simulation framework with a discrete teleport-like action space; thus, no low-level continuous control is considered. To evaluate this work, we migrate their released VLM agent code to our benchmark. 
    
\textbf{VLM-Social-Nav\cite{song_vlm-social-nav_2024}}: a VLM-based navigation system triggered by object detection. VLM model generates high-level action such as ``Move \textlangle DIRECTION\textrangle with \textlangle SPEED\textrangle``, \textlangle DIRECTION\textrangle$\in$\{left, straight, right\}, \textlangle SPEED\textrangle $\in$ \{low down, speed up, constant, stop\}. A discrete candidate action set is then evaluated by a weighted cost function. The action with the lowest cost is selected and published. Since the code has not been released, we re-implement this method in our benchmark. The prompt and workflow of the VLM agent are migrated from the paper.   

To facilitate fair comparison, both our method and baseline methods use GPT-4o in experiments.

\subsection{Quantitative Results}

Table~\ref{tab:results_scenarios} summarizes the comparative results across all scenarios, clearly demonstrating that our approach surpasses both baselines on success rate, collision avoidance, and semantic compliance.

\textbf{Success Rate.}  
In the most challenging ``Follow Doctor'' task (a), only our method achieves a \textbf{100\%} success rate, 
while both baselines completely fail. Across other tasks, our approach maintains 
\textbf{90--100\%} success, whereas baselines typically remain below 60\%.  

\textbf{Collision Rate.}
Our method maintains near-zero collisions in most scenarios, 
achieving \textbf{0\%} in (b), (d), and (e), and remaining low in (a) and (c).  
Baselines show much higher rates, especially in crowded or constrained settings.

\begin{figure*}[t]
    \centering
    \includegraphics[width=0.85\linewidth, clip=true, trim={2cm 6cm 0.5cm 0cm}]{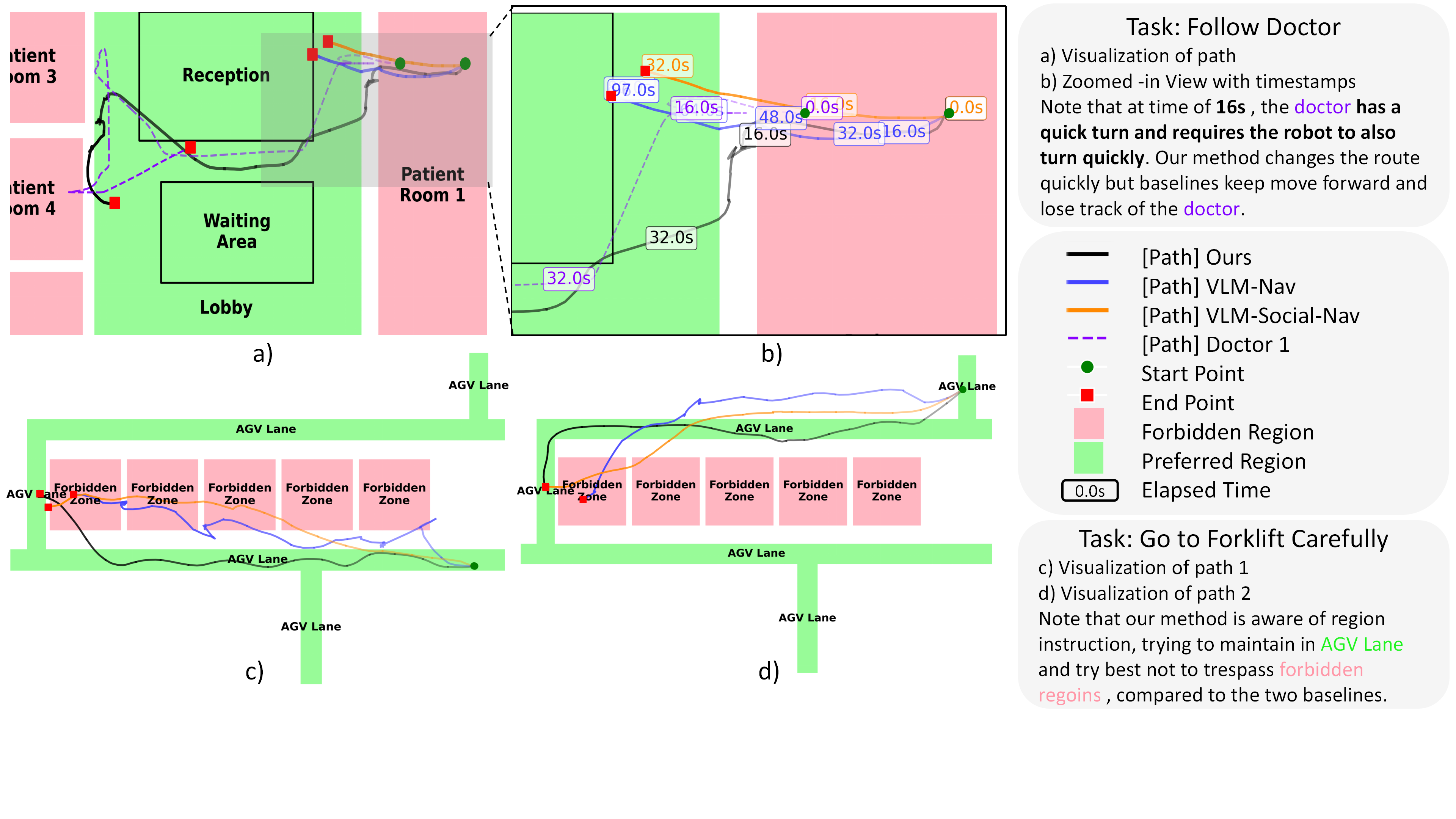}
    \caption{Qualitative Evaluation. This figure explains how our method improves performance compared with the two baselines. In the upper row, our method successfully tracks the moving doctor while two other methods fail due to slow VLM inference and then lose track of the doctor from RGB observation. In the lower row, our method attends to ground line markings much better than the two baselines and abides by the social norm with maximum effort, thanks to extra perception tools. }
    \label{fig:path_visualization}
\end{figure*}

\textbf{Path Smoothness.}  
Our values are generally higher (e.g. \textbf{ 28.85} in (a) vs. \textbf{2.x} for baselines). 


\textbf{Semantic Compliance.}
Our method achieves a higher \emph{Subject Score} in the “Follow Doctor” task (\textbf{78.63}) compared to VLM-Social-Nav (\textbf{71.84}) and VLM-Nav (\textbf{55.53}), indicating better maintenance of socially acceptable distance.
For \emph{Region Score} in the “Go to Forklift Carefully” task, our method (\textbf{80.50}) also outperforms the baselines (VLM-Nav: \textbf{43.28}, VLM-Social-Nav: \textbf{36.56}), demonstrating better adherence to spatial constraints even under VLM latency and occlusions.
These results validate our metric design for assessing both inter-personal distance and spatial rule compliance, both critical for social navigation.


\subsection{System Latency Analysis}

Beyond navigation performance, we further analyze the system latency to demonstrate why disentangling the fast system action generation loop from the VLM inference loop is important. We aim to measure:

\begin{itemize}[leftmargin=*]
    \item \textbf{Slow System (VLM inference)}: 
    from image + instruction input $\rightarrow$ VLM Agent $\rightarrow$ response. 
    For each method, we measure and report the average latency over ten runs.
    \item \textbf{Fast System (Local planning and control)}: 
    from LiDAR input $\rightarrow$ costmap + SFM planner $\rightarrow$ controller $(v, w)$. 
    We report the average latency over 100 runs.
\end{itemize}

\begin{table}[h]
\centering
\caption{Latency Analysis of fast and slow systems.}
\label{tab:latency_results}
\begin{tabular}{lccc}
\toprule
\textbf{Method} & \textbf{System} & \textbf{Avg. Latency (ms)}\\

\midrule
VLM-Nav         & Slow & 9072.10  \\
\midrule
VLM-Social-Nav  & Slow & 1751.80  \\
\midrule
Ours            & Slow & 7094.16  \\
\midrule
SFM controller  & Fast & 5.27  \\
\bottomrule
\end{tabular}
\vspace{-0.3cm}
\end{table}

Table~\ref{tab:latency_results} summarizes the measured latencies of each method. 
We report averaged results in milliseconds (ms).
There is a clear separation in latency: the upper VLM reasoning loop is the main bottleneck, whereas the lower planning loop is efficient ($\sim$ 5ms/step), easily meeting real-time control needs.

 


Our framework trades off latency for better perception results and disentangles low-level action generation from VLM inference. A key example is the most challenging scenario: the “Follow Doctor” task as in Table~\ref{tab:results_scenarios}. Our system achieves a success rate of $100\%$, while both VLM-Nav and VLM-Social-Nav completely fail. This dramatic difference directly stems from the architectural \textbf{decoupling of our fast and slow systems in action generation}, especially in dynamic target interaction: In our method, we project visual markers to the social cost map to associate the geometric obstacle entities on the map with semantic attributes and parameters. Then the map value will be updated by lidar scans with a fast system frequency without further update of the slow system, even with the movement of semantic targets. These results also highlight the general importance of using the fast-slow system in highly dynamic environments when visual-language grounding is required. 


\subsection{Qualitative Results}

We provide two qualitative results in Fig.~\ref{fig:path_visualization} with two case studies.  
The upper row presents an episodic evaluation result of the task \textit{Follow Doctor}.  
Only our method successfully followed the doctor, while both baselines got stuck near the reception desk due to VLM latency, which caused loss of the target.  
The lower row presents an episodic evaluation result of the task \textit{Go to Forklift Carefully}.  
Our agent adapted behavior to different instructions (hurry vs. avoid forbidden zones), while baselines often violated semantic constraints.
Overall, our approach shows stronger robustness and semantic compliance. 



\section{Conclusion and Future Work}

In this paper, we introduced LISN-Bench, a standardized simulation benchmark for language-instructed social navigation that unifies instruction following and scene understanding under continuous, real-time control, filling a critical evaluation gap in socially aware navigation beyond path efficiency and collision avoidance. We further presented Social-Nav-Modulator, a VLM-driven fast–slow architecture that modulates costmaps and SFM controller parameters to translate semantic instructions and social norms into low-dimensional control objectives while preserving the responsiveness and safety properties of classical planners.

Despite these promising results, our work has limitations. The current set of tasks and metrics, while a step forward, is not exhaustive. The evaluation was also conducted purely in simulation. Future work will proceed in three main directions. First, we plan to expand the benchmark with more diverse social contexts, including different cultural norms and more complex multi-person interactions. Second, we will enrich the definition of social norms within the system, moving beyond simple zone-based rules. Finally, we aim to transfer our system to a real-world robotic platform to validate its performance in unstructured human environments.

\def\url#1{}
\bibliographystyle{IEEEtran}
\bibliography{references_zotero}

\begin{thebibliography}{10}
\providecommand{\url}[1]{#1}
\csname url@rmstyle\endcsname
\providecommand{\newblock}{\relax}
\providecommand{\bibinfo}[2]{#2}
\providecommand\BIBentrySTDinterwordspacing{\spaceskip=0pt\relax}
\providecommand\BIBentryALTinterwordstretchfactor{4}
\providecommand\BIBentryALTinterwordspacing{\spaceskip=\fontdimen2\font plus
\BIBentryALTinterwordstretchfactor\fontdimen3\font minus \fontdimen4\font\relax}
\providecommand\BIBforeignlanguage[2]{{%
\expandafter\ifx\csname l@#1\endcsname\relax
\typeout{** WARNING: IEEEtran.bst: No hyphenation pattern has been}%
\typeout{** loaded for the language `#1'. Using the pattern for}%
\typeout{** the default language instead.}%
\else
\language=\csname l@#1\endcsname
\fi
#2}}

\bibitem{feil-seifer_socially_2011}
\BIBentryALTinterwordspacing
D.~Feil-Seifer and M.~J. Matarić, ``Socially {Assistive} {Robotics},'' \emph{IEEE Robotics \& Automation Magazine}, vol.~18, no.~1, pp. 24--31, Mar. 2011.  \url{https://ieeexplore.ieee.org/abstract/document/5751968}
\BIBentrySTDinterwordspacing

\bibitem{song_vlm-social-nav_2024}
\BIBentryALTinterwordspacing
D.~Song, J.~Liang, A.~Payandeh, X.~Xiao, and D.~Manocha, ``\BIBforeignlanguage{en}{{VLM}-social-nav: socially aware robot navigation through scoring using vision-language models},'' July 2024.  \url{http://arxiv.org/abs/2404.00210}
\BIBentrySTDinterwordspacing

\bibitem{kastner_arena_2024}
\BIBentryALTinterwordspacing
L.~Kästner, \emph{et~al.}, ``Arena 3.0: {Advancing} {Social} {Navigation} in {Collaborative} and {Highly} {Dynamic} {Environments},'' June 2024.  \url{http://arxiv.org/abs/2406.00837}
\BIBentrySTDinterwordspacing

\bibitem{tsoi_sean_2022}
\BIBentryALTinterwordspacing
N.~Tsoi, A.~Xiang, P.~Yu, S.~S. Sohn, G.~Schwartz, S.~Ramesh, M.~Hussein, A.~W. Gupta, M.~Kapadia, and M.~Vázquez, ``\BIBforeignlanguage{en}{{SEAN} 2.0: formalizing and generating social situations for robot navigation},'' \emph{\BIBforeignlanguage{en}{IEEE Robotics and Automation Letters}}, vol.~7, no.~4, pp. 11\,047--11\,054, Oct. 2022.  \url{https://ieeexplore.ieee.org/document/9851501/?arnumber=9851501}
\BIBentrySTDinterwordspacing

\bibitem{biswas_socnavbench_2022}
\BIBentryALTinterwordspacing
A.~Biswas, A.~Wang, G.~Silvera, A.~Steinfeld, and H.~Admoni, ``{SocNavBench}: {A} {Grounded} {Simulation} {Testing} {Framework} for {Evaluating} {Social} {Navigation},'' \emph{J. Hum.-Robot Interact.}, vol.~11, no.~3, pp. 26:1--26:24, July 2022.  \url{https://dl.acm.org/doi/10.1145/3476413}
\BIBentrySTDinterwordspacing

\bibitem{perez-higueras_hunavsim_2023}
\BIBentryALTinterwordspacing
N.~Pérez-Higueras, R.~Otero, F.~Caballero, and L.~Merino, ``{HuNavSim}: {A} {ROS} 2 {Human} {Navigation} {Simulator} for {Benchmarking} {Human}-{Aware} {Robot} {Navigation},'' Sept. 2023.  \url{http://arxiv.org/abs/2305.01303}
\BIBentrySTDinterwordspacing

\bibitem{zhang_navid_2024}
\BIBentryALTinterwordspacing
J.~Zhang, K.~Wang, R.~Xu, G.~Zhou, Y.~Hong, X.~Fang, Q.~Wu, Z.~Zhang, and H.~Wang, ``{NaVid}: {Video}-based {VLM} {Plans} the {Next} {Step} for {Vision}-and-{Language} {Navigation},'' June 2024.  \url{http://arxiv.org/abs/2402.15852}
\BIBentrySTDinterwordspacing

\bibitem{cheng_navila_2025}
\BIBentryALTinterwordspacing
A.-C. Cheng, Y.~Ji, Z.~Yang, Z.~Gongye, X.~Zou, J.~Kautz, E.~Bıyık, H.~Yin, S.~Liu, and X.~Wang, ``\BIBforeignlanguage{en}{{NaVILA}: legged robot vision-language-action model for navigation},'' Feb. 2025.  \url{http://arxiv.org/abs/2412.04453}
\BIBentrySTDinterwordspacing

\bibitem{helbing_social_1995}
\BIBentryALTinterwordspacing
D.~Helbing and P.~Molnár, ``\BIBforeignlanguage{en}{Social force model for pedestrian dynamics},'' \emph{\BIBforeignlanguage{en}{Physical Review E}}, vol.~51, no.~5, pp. 4282--4286, May 1995.  \url{https://link.aps.org/doi/10.1103/PhysRevE.51.4282}
\BIBentrySTDinterwordspacing

\bibitem{siciliano_reciprocal_2011}
\BIBentryALTinterwordspacing
J.~Van Den~Berg, S.~J. Guy, M.~Lin, and D.~Manocha, ``Reciprocal n-{Body} {Collision} {Avoidance},'' in \emph{Robotics {Research}}, B.~Siciliano, O.~Khatib, F.~Groen, C.~Pradalier, R.~Siegwart, and G.~Hirzinger, Eds.\hskip 1em plus 0.5em minus 0.4em\relax Berlin, Heidelberg: Springer Berlin Heidelberg, 2011, vol.~70, pp. 3--19.  \url{http://link.springer.com/10.1007/978-3-642-19457-3_1}
\BIBentrySTDinterwordspacing

\bibitem{li_human-aware_2024}
\BIBentryALTinterwordspacing
H.~Li, M.~Li, Z.-Q. Cheng, Y.~Dong, Y.~Zhou, J.-Y. He, Q.~Dai, T.~Mitamura, and A.~G. Hauptmann, ``Human-{Aware} {Vision}-and-{Language} {Navigation}: {Bridging} {Simulation} to {Reality} with {Dynamic} {Human} {Interactions},'' Nov. 2024.  \url{http://arxiv.org/abs/2406.19236}
\BIBentrySTDinterwordspacing

\bibitem{puig_habitat_2023}
\BIBentryALTinterwordspacing
X.~Puig, \emph{et~al.}, ``Habitat 3.0: {A} {Co}-{Habitat} for {Humans}, {Avatars} and {Robots},'' 2023.  \url{https://arxiv.org/abs/2310.13724}
\BIBentrySTDinterwordspacing

\bibitem{khatib_real-time_1986}
\BIBentryALTinterwordspacing
O.~Khatib, ``\BIBforeignlanguage{en}{Real-time obstacle avoidance for manipulators and mobile robots},'' \emph{\BIBforeignlanguage{en}{International Journal of Robotics Research}}, vol.~5, no.~1, pp. 90--98, Mar. 1986.  \url{https://journals.sagepub.com/doi/10.1177/027836498600500106}
\BIBentrySTDinterwordspacing

\bibitem{rosmann_timed-elastic-bands_2015}
\BIBentryALTinterwordspacing
C.~Rösmann, F.~Hoffmann, and T.~Bertram, ``Timed-{Elastic}-{Bands} for time-optimal point-to-point nonlinear model predictive control,'' in \emph{2015 {European} {Control} {Conference} ({ECC})}, July 2015, pp. 3352--3357.  \url{https://ieeexplore.ieee.org/abstract/document/7331052}
\BIBentrySTDinterwordspacing

\bibitem{long_towards_2018}
\BIBentryALTinterwordspacing
P.~Long, T.~Fan, X.~Liao, W.~Liu, H.~Zhang, and J.~Pan, ``\BIBforeignlanguage{en}{Towards optimally decentralized multi-robot collision avoidance via deep reinforcement learning},'' in \emph{\BIBforeignlanguage{en}{2018 {IEEE} {International} {Conference} on {Robotics} and {Automation} ({ICRA})}}, May 2018, pp. 6252--6259.  \url{https://ieeexplore.ieee.org/abstract/document/8461113}
\BIBentrySTDinterwordspacing

\bibitem{li_sarl_2019}
\BIBentryALTinterwordspacing
K.~Li, Y.~Xu, J.~Wang, and M.~Q.-H. Meng, ``\BIBforeignlanguage{en}{{SARL}: deep reinforcement learning based human-aware navigation for mobile robot in indoor environments},'' in \emph{\BIBforeignlanguage{en}{2019 {IEEE} {International} {Conference} on {Robotics} and {Biomimetics} ({ROBIO})}}, Dec. 2019, pp. 688--694.  \url{https://ieeexplore.ieee.org/abstract/document/8961764}
\BIBentrySTDinterwordspacing

\bibitem{chen_crowd-robot_2019}
\BIBentryALTinterwordspacing
C.~Chen, Y.~Liu, S.~Kreiss, and A.~Alahi, ``\BIBforeignlanguage{en}{Crowd-robot interaction: crowd-aware robot navigation with attention-based deep reinforcement learning},'' in \emph{\BIBforeignlanguage{en}{2019 {International} {Conference} on {Robotics} and {Automation} ({ICRA})}}, May 2019, pp. 6015--6022.  \url{https://ieeexplore.ieee.org/abstract/document/8794134}
\BIBentrySTDinterwordspacing

\bibitem{yao_sonic_2025}
\BIBentryALTinterwordspacing
J.~Yao, X.~Zhang, Y.~Xia, Z.~Wang, A.~K. Roy-Chowdhury, and J.~Li, ``\BIBforeignlanguage{en}{{SoNIC}: safe social navigation with adaptive conformal inference and constrained reinforcement learning},'' Feb. 2025.  \url{http://arxiv.org/abs/2407.17460}
\BIBentrySTDinterwordspacing

\bibitem{martini_adaptive_2024}
\BIBentryALTinterwordspacing
M.~Martini, N.~Pérez-Higueras, A.~Ostuni, M.~Chiaberge, F.~Caballero, and L.~Merino, ``\BIBforeignlanguage{en}{Adaptive social force window planner with reinforcement learning},'' in \emph{\BIBforeignlanguage{en}{2024 {IEEE}/{RSJ} {International} {Conference} on {Intelligent} {Robots} and {Systems} ({IROS})}}, Oct. 2024, pp. 4816--4822.  \url{https://ieeexplore.ieee.org/abstract/document/10802383}
\BIBentrySTDinterwordspacing

\bibitem{gupta_social_2018}
A.~Gupta, J.~Johnson, L.~Fei-Fei, S.~Savarese, and A.~Alahi, ``Social gan: {Socially} acceptable trajectories with generative adversarial networks,'' in \emph{Proceedings of the {IEEE} conference on computer vision and pattern recognition}, 2018, pp. 2255--2264.

\bibitem{shah_vint_2023}
\BIBentryALTinterwordspacing
D.~Shah, A.~Sridhar, N.~Dashora, K.~Stachowicz, K.~Black, N.~Hirose, and S.~Levine, ``{ViNT}: {A} {Foundation} {Model} for {Visual} {Navigation},'' Oct. 2023.  \url{http://arxiv.org/abs/2306.14846}
\BIBentrySTDinterwordspacing

\bibitem{goetting_end--end_2024}
\BIBentryALTinterwordspacing
D.~Goetting, H.~G. Singh, and A.~Loquercio, ``\BIBforeignlanguage{en}{End-to-end navigation with vision language models: transforming spatial reasoning into question-answering},'' Nov. 2024.  \url{http://arxiv.org/abs/2411.05755}
\BIBentrySTDinterwordspacing

\bibitem{shah_lm-nav_2023}
\BIBentryALTinterwordspacing
D.~Shah, B.~Osiński, and S.~Levine, ``\BIBforeignlanguage{en}{Lm-nav: robotic navigation with large pre-trained models of language, vision, and action},'' in \emph{\BIBforeignlanguage{en}{Conference on {Robot} {Learning}}}.\hskip 1em plus 0.5em minus 0.4em\relax PMLR, 2023, pp. 492--504.  \url{https://proceedings.mlr.press/v205/shah23b}
\BIBentrySTDinterwordspacing

\bibitem{chen_how_2023}
\BIBentryALTinterwordspacing
J.~Chen, G.~Li, S.~Kumar, B.~Ghanem, and F.~Yu, ``How {To} {Not} {Train} {Your} {Dragon}: {Training}-free {Embodied} {Object} {Goal} {Navigation} with {Semantic} {Frontiers},'' May 2023.  \url{http://arxiv.org/abs/2305.16925}
\BIBentrySTDinterwordspacing

\bibitem{elnoor_vi-lad_2025}
\BIBentryALTinterwordspacing
M.~Elnoor, K.~Weerakoon, G.~Seneviratne, J.~Liang, V.~Rajagopal, and D.~Manocha, ``Vi-{LAD}: {Vision}-{Language} {Attention} {Distillation} for {Socially}-{Aware} {Robot} {Navigation} in {Dynamic} {Environments},'' Mar. 2025.  \url{http://arxiv.org/abs/2503.09820}
\BIBentrySTDinterwordspacing

\bibitem{sathyamoorthy_convoi_2024}
\BIBentryALTinterwordspacing
A.~J. Sathyamoorthy, K.~Weerakoon, M.~Elnoor, A.~Zore, B.~Ichter, F.~Xia, J.~Tan, W.~Yu, and D.~Manocha, ``Convoi: {Context}-aware navigation using vision language models in outdoor and indoor environments,'' in \emph{2024 {IEEE}/{RSJ} {International} {Conference} on {Intelligent} {Robots} and {Systems} ({IROS})}.\hskip 1em plus 0.5em minus 0.4em\relax IEEE, 2024, pp. 13\,837--13\,844.  \url{https://ieeexplore.ieee.org/abstract/document/10802716/}
\BIBentrySTDinterwordspacing

\bibitem{shcherbyna1_arena_2024}
\BIBentryALTinterwordspacing
V.~Shcherbyna1, L.~Kästner, D.~Diaz, H.~G. Nguyen, M.~H.-K. Schreff, T.~Lenz, J.~Kreutz, A.~Martban, H.~Zeng, and H.~Soh, ``Arena 4.0: {A} {Comprehensive} {ROS2} {Development} and {Benchmarking} {Platform} for {Human}-centric {Navigation} {Using} {Generative}-{Model}-based {Environment} {Generation},'' Sept. 2024.  \url{http://arxiv.org/abs/2409.12471}
\BIBentrySTDinterwordspacing

\bibitem{haarnoja_soft_2018}
\BIBentryALTinterwordspacing
T.~Haarnoja, A.~Zhou, P.~Abbeel, and S.~Levine, ``\BIBforeignlanguage{en}{Soft actor-critic: off-policy maximum entropy deep reinforcement learning with a stochastic actor},'' in \emph{\BIBforeignlanguage{en}{Proceedings of the 35th {International} {Conference} on {Machine} {Learning}}}.\hskip 1em plus 0.5em minus 0.4em\relax PMLR, July 2018, pp. 1861--1870.  \url{https://proceedings.mlr.press/v80/haarnoja18b.html}
\BIBentrySTDinterwordspacing

\bibitem{yuan_robopoint_2024}
\BIBentryALTinterwordspacing
W.~Yuan, J.~Duan, V.~Blukis, W.~Pumacay, R.~Krishna, A.~Murali, A.~Mousavian, and D.~Fox, ``\BIBforeignlanguage{en}{{RoboPoint}: a vision-language model for spatial affordance prediction for robotics},'' June 2024.  \url{http://arxiv.org/abs/2406.10721}
\BIBentrySTDinterwordspacing

\bibitem{ren_grounded_2024}
\BIBentryALTinterwordspacing
T.~Ren, \emph{et~al.}, ``\BIBforeignlanguage{en}{Grounded {SAM}: assembling open-world models for diverse visual tasks},'' Jan. 2024.  \url{http://arxiv.org/abs/2401.14159}
\BIBentrySTDinterwordspacing

\bibitem{openai_gpt-4o_2024}
\BIBentryALTinterwordspacing
OpenAI, \emph{et~al.}, ``\BIBforeignlanguage{en}{{GPT}-4o system card},'' Oct. 2024.  \url{http://arxiv.org/abs/2410.21276}
\BIBentrySTDinterwordspacing

\end{thebibliography}

\end{document}